\documentclass[mathpazo]{cicp}

\usepackage{amsmath,amsfonts,amsthm}
\usepackage{booktabs}
\usepackage{longtable}
\usepackage{setspace}
\usepackage{amssymb}
\usepackage{fancyhdr}
\usepackage{mathrsfs}
\usepackage{amsfonts}
\usepackage{cases}
\usepackage{verbatim}
\usepackage{graphicx}      
\usepackage{epstopdf}
\usepackage{grffile}
\graphicspath{{figures/}}
\usepackage{subfig}
\usepackage{caption}
\usepackage{tikz}
\usepackage[ruled,boxed]{algorithm2e}

\providecommand{\keywords}[1]{\textbf{\textit{Index terms---}} #1}

\begin{document}
	
\title{A Novel Adaptive Causal Sampling Method for Physics-Informed Neural Networks}

 \author[Guo J et.~al.]{Jia Guo\affil{1}, Haifeng Wang\affil{1}
        and Chenping Hou\affil{1}\comma\corrauth}
 \address{\affilnum{1}\ College of Science,
          National University of Defense Technology,
          Changsha 410073, P.R. China.}
 \emails{{\tt guojia14@nudt.edu.cn} (J.~Guo),
 		 {\tt wanghaifeng20@nudt.edu.cn} (H.~Wang),
          {\tt hcpnudt@hotmail.com} (C.~Hou)}

\begin{abstract}
Physics-Informed Neural Networks (PINNs) have become a kind of attractive machine learning method for obtaining solutions of partial differential equations (PDEs). Training PINNs can be seen as a semi-supervised learning task, in which only exact values of initial and boundary points can be obtained in solving forward problems, and in the whole spatio-temporal domain collocation points are sampled without exact labels, which brings training difficulties. Thus the selection of collocation points and sampling methods are quite crucial in training PINNs. Existing sampling methods include fixed and dynamic types, and in the more popular latter one, sampling is usually controlled by PDE residual loss. We point out that it is not sufficient to only consider the residual loss in adaptive sampling and sampling should obey temporal causality. We further introduce temporal causality into adaptive sampling and propose a novel adaptive causal sampling method to improve the performance and efficiency of PINNs. Numerical experiments of several PDEs with high-order derivatives and strong nonlinearity, including Cahn Hilliard and KdV equations, show that the proposed sampling method can improve the performance of PINNs with few collocation points. We demonstrate that by utilizing such a relatively simple sampling method, prediction performance can be improved up to two orders of magnitude compared with state-of-the-art results with almost no extra computation cost, especially when points are limited.


\end{abstract}


\maketitle


\section{Introduction}
\label{Introduction}
Many natural phenomena and physics laws can be described by partial differential equations (PDEs), which are powerful modeling tools in quantum mechanics, fluid dynamics and phase field, etc. Traditional numerical methods play an important role in solving many kinds of PDEs in the science and engineering fields. However, complex nonlinear problems always require the delicate design of schemes, heavy preparation for mesh generation, and expensive computational costs. Due to the rapid development of machine learning, data-driven methods attract much attention not only in traditional areas of the computer field, such as computer vision (CV) and natural language processing (NLP) but also scientific computing field, which motivates a new field of scientific machine learning (SciML)\cite{PIML}\cite{machinelearning}\cite{Carleo_2017}\cite{Arridge2019SolvingIP}\cite{Karpatne2017TheoryGuidedDS}. Data-driven methods for solving PDEs utilize machine learning tools to learn the nonlinear mapping from inputs (spatio-temporal data) to outputs (solution of PDEs), which omits heavy preparation and improves computational efficiency. Physics-Informed Neural Networks (PINNs) \cite{PINN2019} are a kind of representative work in this field. They have received extensive attention and much recent work\cite{activation} \cite{domain} \cite{Yang_2021} \cite{fPINNs} \cite{CiCP-28-2002} based on PINNs are put forward immediately after them.

PINNs are a class of machine learning algorithms where the loss function is specially designed based on the given PDEs with initial and boundary conditions. Automatic-differentiation technique \cite{autodiff} is utilized in PINNs to calculate the exact derivatives of the variables. By informing physics prior information into machine learning method, PINNs enhance interpretability and thus are not a class of pure black-box models. According to the classification of machine learning, PINNs can be seen as semi-supervised learning algorithms. PINNs are trained to not only minimize the mean squared error between the prediction of initial and boundary points and their given exact values but also satisfy PDEs in collocation points. The former is easy to be implemented, while the latter needs to be explored further. Therefore the selection and sampling of collocation points are vital for the prediction and efficiency of PINNs. The traditional sampling method of PINNs is to sample uniform or random collocation points before training, which is a kind of fixed sampling method. Then several adaptive sampling methods have been proposed, including RAR\cite{deepxde}, adaptive sampling\cite{solving}, bc-PINN method\cite{sequential}, importance sampling\cite{importancesampling}, RANG\cite{RANG}, RAD and RAR-D\cite{comprehensive}, Evo and Causal Evo\cite{Rethinking}.

Though the importance of sampling for PINNs has been enhanced to a certain extent in these works, temporal causality has not been emphasized in sampling, especially for solving time-dependent PDEs. Wang et al. \cite{Respectingcausality} proposed the causal training algorithm for PINNs by informing designed residual-based temporal causal weights into the loss function. This algorithm can make sure that loss in the previous time should be minimized in advance, which respects temporal causality. However, in \cite{Respectingcausality}, the collocation points are sampled evenly and fixedly in each spatio-temporal sub-domain, which is not suitable in many situations. We indicate that collocation points should also be sampled under the foundation of respecting temporal causality. This argument stems from traditional numerical schemes. Specifically, the designed iterative schemes calculate the solution from the initial moment to the next moment according to the time step. Similar to this, the sampling method should also obey this temporal causality guideline.

Motivated by traditional numerical schemes and temporal causality, we propose a novel adaptive causal sampling method that collocation points are adaptively sampled according to both PDE residual loss and temporal causal weight. This idea is original from the adaptation mechanism of the finite element method (FEM), which can better improve computational efficiency, and from the temporal causality of traditional numerical schemes, which obeys the temporal order.

In this paper, we mainly focus on the sampling methods for PINNs. Our specific contributions are as follows:

\begin{itemize}
	\item We analyze the failure of adaptive sampling and figure out that sampling should obey temporal causality, otherwise leading to sampling confusion and trivial solution of PINNs.
	\item We introduce temporal causality into sampling and propose a novel adaptive causal sampling method.
	\item We investigate our proposed method in several numerical experiments and gain better results than state-of-the-art sampling methods with almost no extra computational cost and with few points, which shows the high efficiency of our proposed method and potential in computationally complex problems, such as large-scale problems and high-dimensional problems.
\end{itemize}

The structure of this paper is organized as follows. Section \ref{sec2} briefly introduces the main procedure of PINNs and analyzes the necessity of introducing temporal causality in sampling. Section \ref{sec3} proposes a novel adaptive causal sampling method. In Section \ref{sec4}, we investigates several numerical experiments to demonstrate the performance of proposed method. The final section concludes the whole paper.

\section{Background}
\label{sec2}
In this section, we first provide a brief overview of PINNs. Then, by investigating an illustrative example, we analyze the necessity of temporal causality in sampling.
\subsection{Physics-Informed Neural Networks}\label{sec2.1}
PINNs are a class of machine learning algorithms where the prior physics information including initial and boundary conditions, and corresponding PDEs form are informed into the loss function. Here we consider the general form of nonlinear PDEs with initial and boundary conditions
\begin{equation}\label{pde}
\begin{aligned}
&u_t+\mathcal{N}[u]=0,t\in[0,T],x\in\Omega,\\
&u(0,x)=g(x),x\in\Omega,\\
&\mathcal{B}[u]=0,t\in[0,T],x\in\partial\Omega,
\end{aligned}
\end{equation}
where $x$ and $t$ are the space and time coordinates respectively, $u$ is the solution of PDEs system of Equation \eqref{pde}. $\Omega$ is the computational domain and $\partial \Omega$ represents the boundary. $\mathcal{N}[.]$ is a nonlinear differential operator, $g(x)$ is the initial function, and $\mathcal{B}[.]$ represents boundary operator, including periodic, Dirichlet, Neumann boundary conditions etc.

The universal approximation theory\cite{Universal} can guarantee that, there exists a deep neural network $\hat{u}_{\theta}(t,x)$ with nonlinear activation function $\sigma$ and tunable parameters $\theta$ namely weights and bias, such that the PDEs solution $u(t,x)$ can be approximated by it. Then the prediction of the solution $u(t,x)$ can be converted to the optimization problem of training a deep learning model. The aim of training PINNs is to minimize the loss function and find the optimal parameters $\theta$. The usual form of loss function in PINNs is composited by three parts with tunable coefficients
\begin{equation}\label{loss}
\mathcal{L}(\theta)=\lambda_{ic}\mathcal{L}_{ic}(\theta)+\lambda_{bc}\mathcal{L}_{bc}(\theta)+\lambda_{res}\mathcal{L}_{res}(\theta)
\end{equation}
where
\begin{equation}\label{lossall}
\begin{aligned}
&\mathcal{L}_{ic}(\theta)=\frac{1}{N_{ic}}\sum_{i=1}^{N_{ic}}|\hat{u}_{\theta}(0,x_{ic}^i)-g(x_{ic}^i)|^2,\\
&\mathcal{L}_{bc}(\theta)=\frac{1}{N_{bc}}\sum_{i=1}^{N_{bc}}|\mathcal{B}[\hat{u}_{\theta}](t_{bc}^i,x_{bc}^i)|^2,\\
&\mathcal{L}_{res}(\theta)=\frac{1}{N_{res}}\sum_{i=1}^{N_{res}}|\frac{\partial{\hat{u}_{\theta}}}{\partial t}(t_{res}^i,x_{res}^i)+\mathcal{N}[\hat{u}_{\theta}](t_{res}^i,x_{res}^i)|^2.
\end{aligned}
\end{equation}
$\{0,x_{ic}^i\}_{i=1}^{N_{ic}}$, $\{t_{bc}^i,x_{bc}^i\}_{i=1}^{N_{bc}}$ and $\{t_{res}^i,x_{res}^i\}_{i=1}^{N_{res}}$ are initial data, boundary data and residual collocation points respectively, which are the inputs of PINNs. In the loss function, the calculation of derivatives, e.g. $\frac{\partial{\hat{u}_{\theta}}}{{\partial t}}$ can be obtained via automatic differentiation\cite{autodiff}. Besides, the gradients with respect to network parameters $\theta$ are also computed via this technique. Moreover, the hyper-parameters ${\lambda_{ic},\lambda_{bc},\lambda_{res}}$ are usually tuned by users or by automatic algorithms in order to balance training of different loss terms. 

\subsection{Analysis of sampling}\label{sec2.2}
\subsubsection{Existing sampling methods}\label{sec2.2.1}

In original PINNs\cite{PINN2019}, collocation points are uniformly or randomly sampled before the whole training procedure, which can be seen as a fixed type of sampling. This type is quite useful for solving some PDEs, however, for more complicated PDEs, it has difficulties in both prediction accuracy and convergence efficiency. To improve the sampling performance for PINNs, the adaptive idea is utilized in PINNs, which forms the adaptive type of sampling. Residual-based adaptive refinement (RAR) in PINNs is first proposed by Lu\cite{deepxde} which adds new collocation points in areas with large PDE residuals. This kind of adaptive sampling methods pursue improved prediction accuracy by automatically sampling more collocation points. 

Compared with these sampling methods with increasing number of sampled points, we aim to achieve better accuracy under the foundation of limited and few points, which can save computing resources. Thus in this paper we focus on automatically adaptive sampling methods with limited sampled points. First we analyze the failure of existing adaptive sampling methods, then propose our novel sampling method to overcome the failure. 

\subsubsection{An illustrative example}\label{sec2.2.2}
To motivate our method in Section \ref{sec3}, here we provide an illustrative example with the adaptive sampling method. We consider the one-dimensional Cahn Hilliard equation with initial and boundary conditions
\begin{equation}\label{ch}
\begin{aligned}
&u_t-\nabla^2(u^3-u-0.02\nabla^2{u})=0, (x,t)\in \Omega \times \left (0,T\right],\\
&u(x,0)=\cos(\pi x)-\exp{(-4(\pi x)^2)}, x\in \Omega,\\
&u(-x,t)= u(x,t), (x,t)\in \partial\Omega \times \left(0,T\right],\\
&u_x(-x,t)= u_x(x,t), (x,t)\in \partial\Omega \times \left(0,T\right],\\
\end{aligned}
\end{equation}
where $\Omega$ represents the whole computational space and $\partial\Omega$ describes the boundary.

This initial boundary value problem (IBVP) is quite difficult to be solved with the original PINNs\cite{PINN2019}. Mattey has investigated this example by original PINNs and proposed backward compatible sequential PINN method (bc-PINN)\cite{sequential} to improve the performance. In \cite{sequential}, $2\times10^6$ collocation points are sampled in both original PINNs and bc-PINN which is a quite numerous number. The relative $L^2$ error obtained by bc-PINN solution is 0.0186 whereas for the original PINNs solution the error is 0.8594. 

We utilize causal PINN\cite{Respectingcausality}, a mature PINNs training method, with the neural network of 4 hidden layers with 128 neurons in each layer. We divide the whole computational domain into $N_t$ spatio-temporal sub-domains. The division is even in time with expression as $0=t_1<t_2<...<t_{N_t}=T$ and we record $i$-th time interval $[t_{i},t_{i+1})$ as $T_i$. Suppose that in $i$-th sub-domain, $N_i$ collocation points $\{t_r^i,x_r^i\}_{i=1}^{N_{i}}$ are sampled. Then we remark temporal PDE residual loss of $i$-th sub-domain as $\mathcal{L}_{res}(T_i,\theta)$ and the calculation is as follows

\begin{equation}\label{equ:losst}
\begin{aligned}
\mathcal{L}_{res}(T_i,\theta)&=\frac{1}{N_{i}}\sum_{i=1}^{N_{i}}|\frac{\partial{u_{\theta}}}{\partial t}(t_{res}^i,x_{res}^i)+\nabla^2({{u_{\theta}}^3(t_{res}^i,x_{res}^i)}-u_{\theta}(t_{res}^i,x_{res}^i)-0.02\nabla^2{u_{\theta}(t_{res}^i,x_{res}^i)})|^2.
\end{aligned}
\end{equation}

In order to investigate the adaptive method with limited points, we consider another residual-based adaptive sampling\cite{importancesampling} that collocation points are resampled according to the distribution proportional to the PDE residual loss. Specifically, collocation points in each spatio-temporal sub-domain are sampled according to the distribution proportional $ratio(i)$ 

\begin{equation}\label{adp}
ratio(i)=\mathcal{L}_{res}(T_i,\theta)/\sum_{i=1}^{N_t}{\mathcal{L}_{res}(T_i,\theta)}
\end{equation}
which is related to the residual loss in each spatio-temporal sub-domain. The total number of automatic sampled collocation points stays unchanged 1000 in this example. Here we use Latin Hypercube Sampling strategy (LHS) \cite{LHS} to generate 1000 collocation points before the causal training procedure. Then we recalculate the distribution proportional ratio and resample points every 1000 iterations of gradient descent algorithm, which can be seen as a dynamic type of sampling. We train the causal PINN model via Adam optimizer for $1\times10^5$ iterations. 

\begin{figure}[h!]
	\flushleft
	\includegraphics[width=1.0\textwidth]{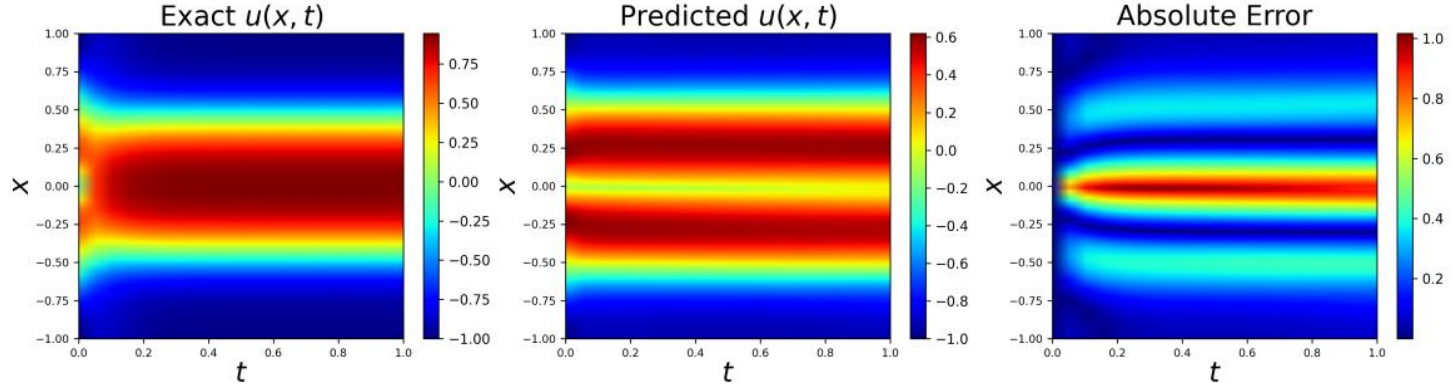}
	\caption{Cahn Hillard equation with adaptive sampling: Reference solution, prediction solution, and error results. The relative $L^2$ error is $4.684e^{-1}$.}
	\label{fig:ch2_adap}
\end{figure} 

\begin{figure}[htbp]
	\centering
	\includegraphics[width=1.0\textwidth]{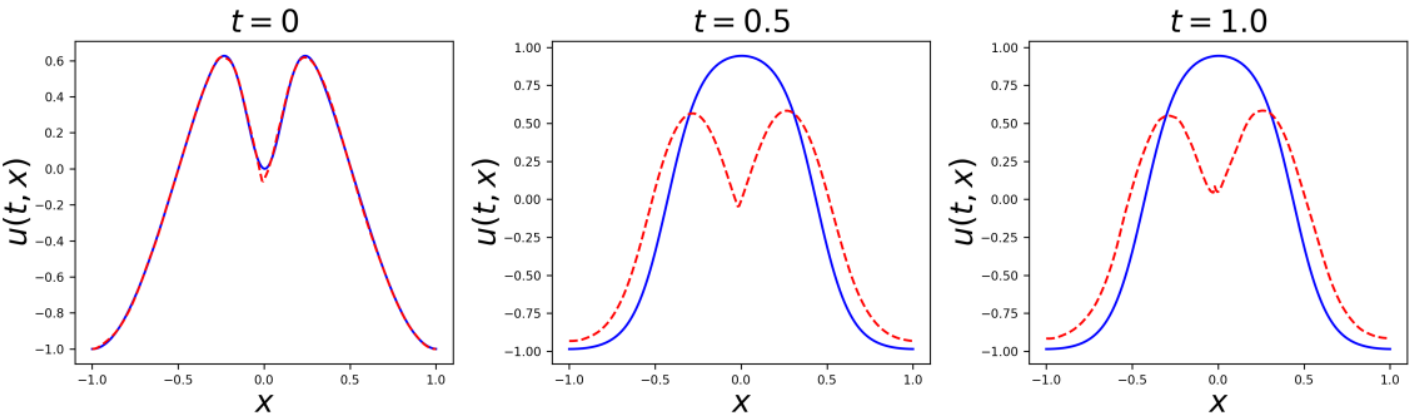}
	\caption{Cahn Hillard equation with adaptive sampling: Result Snapshots at different time.}
	\label{fig:ch2_adap_snap}
\end{figure}

After such sufficient training, the whole loss is reduced to $10^{-3}$ magnitude, which means PINNs converge. However, the prediction results in Figure \ref{fig:ch2_adap} and \ref{fig:ch2_adap_snap} show that such adaptive sampling method leads to convergence to wrong solution. 

\subsubsection{Analyze the failure via the sampling perspective}\label{sec2.2.3}
In Section \ref{sec2.2.2}, the adaptive sampling suffers in converging to correct solution. In this section, we analyze this failure in the sampling perspective, and we first briefly introduce the specially designed residual loss in causal PINN:
\begin{equation}\label{wloss}
\begin{aligned}
\mathcal{L}_{res}(\theta)=\frac{1}{N_t}\sum_{i=1}^{N_t}\omega_i\mathcal{L}_{res}(T_i,\theta),
\end{aligned}
\end{equation}
which consists of temporal PDE residual loss in each sub-domain in Equation \eqref{equ:losst} and the corresponding temporal weight

\begin{equation}\label{tweight}
\omega_i=\exp(-\epsilon\sum_{k=1}^{i-1}\mathcal{L}_{res}(T_k,\theta)).
\end{equation}
Here $\epsilon$ is a tunable hyper-parameter that determines the "slope" of temporal weights. In this paper, we choose its value from $\{10^{-2},10^{-1},10^{0},10^{1},10^{2}\}$. The intuitive explanation of temporal weight in Equation \eqref{tweight} is the temporal priority of training, which represents temporal training order. Suppose that in $i$-th sub-domain, the value of temporal weight is large, which represents that the residual loss should be minimized in advance. When the value is zero, it means that the corresponding residual loss has not been minimized yet.

\begin{figure}[h!]
	\centering
	\includegraphics[width=1.0\textwidth]{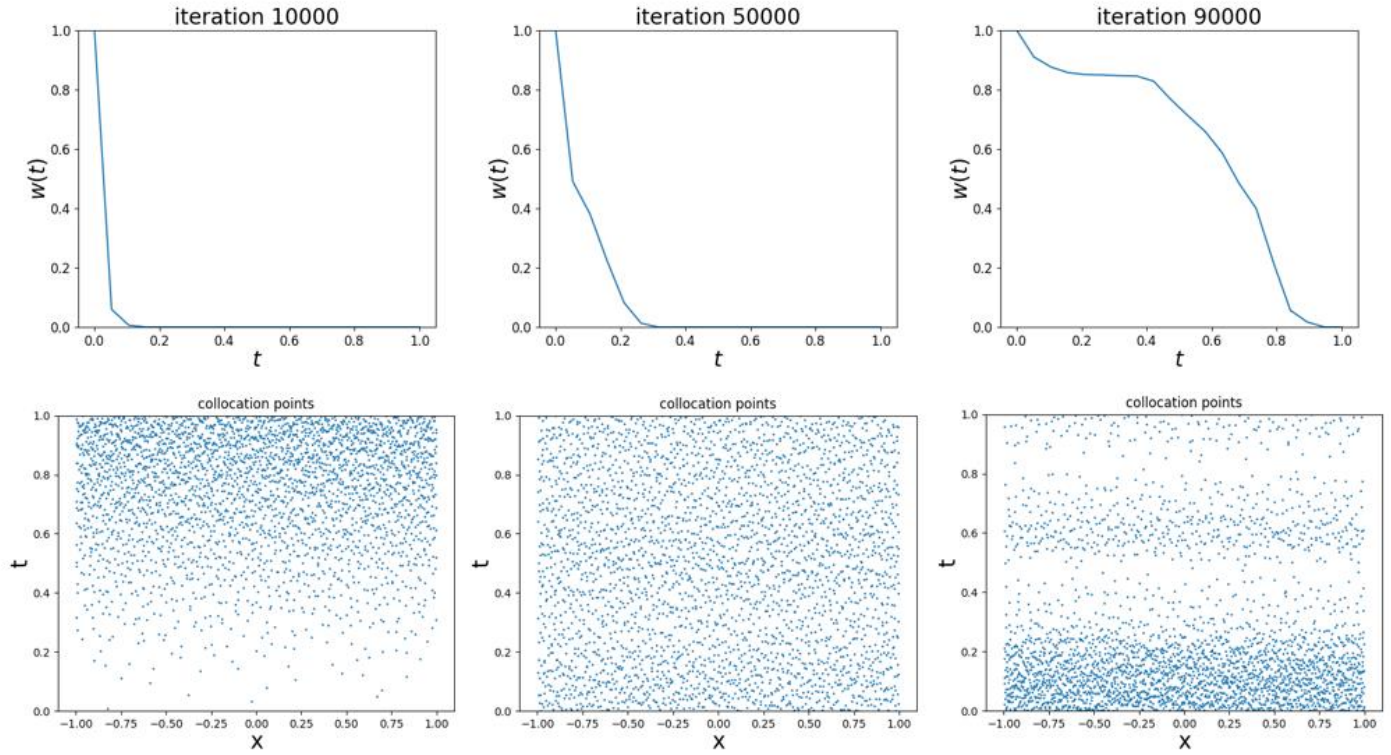}
	\caption{Illustrative example: Top: Temporal weights $w(t)$ at different training iterations. Bottom: Distribution of collocation points in the whole spatio-temporal domain at corresponding iteration with adaptive sampling.}
	\label{fig:ch2_adap_weight&points}
\end{figure}

Figure \ref{fig:ch2_adap_weight&points} shows the values of temporal weights in the whole temporal domain and the distribution of sampled collocation points at different training iterations. The top results show the temporal weights $\omega(t)$ in different training iterations, which illustrates the training process. For clearer illustration, we take the left figure in the first row as an example. At $1\times 10^4$ training iteration, the model in $(x,t)\in[-1,1]\times [0,0.1]$ sub-domain is trained, however in the remaining sub-domains where temporal weights are zero, it has not been trained yet. 

In the bottom figures, we can see that at the beginning of training, collocation points gather at the area near $t=1$, namely most sampled points are far away from the initial time, where temporal weights are large. Then as training iteration increases, points tend to gather at the area near $t=0$, which is very unreasonable and breaks the rule of temporal causality order. Thus we point out that the adaptive sampling in which collocation points are sampled only based on the residual loss is not reliable.

\subsubsection{Sampling should obey temporal causality}
In order to explore the reason for above failure and propose a reliable sampling method, we analyze the residual loss $L_{res}(t_i,\theta)$ and the temporal causality mechanics in causal PINN. According to Equation \eqref{lossall} and \eqref{equ:losst}, we have
\begin{equation}\label{lt}
\begin{aligned}
\mathcal{L}_{res}(T_i,\theta)=\frac{1}{N_{i}}\sum_{i=1}^{N_{i}}|\frac{\partial{\hat{u}_{\theta}}}{\partial t}(t_{res}^i,x_{res}^i)+\mathcal{N}[\hat{u}_{\theta}](t_{res}^i,x_{res}^i)|^2.
\end{aligned}
\end{equation}
Then we utilize the forward Euler scheme 
\begin{equation}\label{Euler}
\begin{aligned}
\frac{\partial{\hat{u}_{\theta}}}{{\partial t}}=\frac{\hat{u}_{\theta}(t_{i},x)-\hat{u}_{\theta}(t_{i-1},x)}{\Delta t}+O({\Delta t})
\end{aligned}
\end{equation}
with $\Delta t$ as the time step to discrete the partial derivative $\frac{\partial{\hat{u}_{\theta}}}{{\partial t}}$, then Equation \eqref{lt} can be transformed to
\begin{equation}\label{lt1}
\begin{aligned}
\mathcal{L}_{res}(T_i,\theta) &\approx \frac{1}{N_{i}}\sum_{j=1}^{N_{i}}|\frac{\hat{u}_{\theta}(t_{res}^i,x_{res}^j)-\hat{u}_{\theta}(t_{res}^{i-1},x_{res}^j)}{\Delta t} +\mathcal{N}[\hat{u}_{\theta}](t_{res}^i,x_{res}^i)|^2\\
&\approx \frac{|\Omega|}{\Delta t^2} \int_{\Omega} |\hat{u}_{\theta}(t_{res}^i,x_{res}^j)-\hat{u}_{\theta}(t_{res}^{i-1},x_{res}^j)+\Delta t \mathcal{N}[\hat{u}_{\theta}](t_{res}^i,x_{res}^i)|^2 dx.
\end{aligned}
\end{equation}

We can see that the foundation of minimization of temporal residual loss $\mathcal{L}_{res}(T_i,\theta)$ is the accurate prediction of both $\hat{u}_{\theta}(t_{res}^i,x_{res}^j)$ and $\hat{u}_{\theta}(t_{res}^{i-1},x_{res}^j)$, which requires sufficient training in $[t_{i-1},t_i)\times \Omega$ with enough collocation points. Thus in the scenario of limited points, collocation points need to be sampled where training priority needs most, in order to guarantee the minimization is carried out correctly and reliably. 

Then according to the temporal causality mechanics in Equation \eqref{wloss} and \eqref{tweight}, when $\mathcal{L}_{res}(T_i,\theta)$ is minimized, then temporal residual loss $\mathcal{L}_{res}(T_{i+1},\theta)$ can begin to be minimized, which constitutes the training order of PINN. Analogy to traditional iterative schemes\cite{cfd}, the calculation is done from the initial time to later one according to the time step, respecting the temporal causality. Only when the solution in front time is calculated well the solution in later time can be predicted accurately. Based on the above analysis, we point out that sampling should also respect this principle to ensure that loss is minimized on the premise that the solution converges to the right one. 

Above adaptive sampling method obviously disobey this principle. Specifically, at the beginning of training, very less collocation points near initial time are sampled, due to the high residual loss concentrating on latter area, which cannot ensure adequate training in front area. Thus prediction solution of causal PINN tends to converge to trivial one. After training for several iterations, residual loss in front area is quite high where more points are sampled, however this sampling behavior breaks the temporal causality principle. In this section, we analyze the failure of adaptive sampling and propose our argument: sampling method should obey temporal causality. And in Section \ref{sec3} and \ref{sec4}, we will further verify this by proposing new sampling method and exploring in numerical experiments.

\section{Adaptive Causal Sampling method}\label{sec3}

Based on the above illustrative example and simple analysis, we consider introducing temporal causality into the sampling method. First, we divide the whole computational domain into $N_t$ spatio-temporal sub-domains along the direction of time. We fix the total amount of sampled collocation points as unchanged $N_r$ during the whole training procedure. The amount of collocation points sampled in each spatio-temporal sub-domain is calculated by the distribution ratio, which is adaptively related to both PDE residual loss and temporal weight in each sub-domains. Here, we propose the novel temporal causal sampling distribution ratio:

\begin{equation}\label{rat}
ratio(i) = \omega_i*\mathcal{L}_{res}(T_i,\theta)/\sum_{i=1}^{N_t}{\omega_i*\mathcal{L}_{res}(T_i,\theta)},
\end{equation}
where $\mathcal{L}_{res}(T_i,\theta)$ is calculated by Equation \eqref{equ:losst}
and temporal weight $\omega_i$ is calculated by Equation \eqref{tweight}.


In Equation \eqref{adp}, the existing residual-based adaptive sampling only focus on PDE residual loss. The intuition is that where PDE residual loss is larger, more collocation points should be sampled in order to minimize the loss faster. However, in Section \ref{sec2.2.3} we point out that this existing adaptive sampling is not reliable. Then we introduce temporal weight into the distribution ratio in Equation \eqref{rat}, thus sampling can be controlled by both PDE residual loss and temporal causality. 

Due to the total number of points is limited, points should be sampled where needed most. According to the temporal causal sampling distribution ratio, when both temporal weight value and PDE residual loss are large, loss in corresponding sub-domains needs to be minimized in advance, where more points should be sampled to accelerate the minimization procedure, thus the value of ratio should be large. And when in some sub-domains the value of temporal weight is very small or nearly zero, which means the temporal order of training is relatively low, even though PDE residual loss is quite large, the ratio in these areas should be small. Besides, the value of temporal weight dynamically changes according to the residual loss of previous sub-domains, thus the sampled points dynamically and adaptively change from front spatio-temporal sub-domains to latter ones, which obeys the temporal causality.

\begin{algorithm}[h!]
	\caption{Resampling procedure}\label{algorithm1}
	\label{alg:resample}
	\KwIn{Amount of all sampling points $N_r$, temporal residual loss $\mathcal{L}_{res}(T_i,\theta)$ and temporal causal weight $W_i$;}
	\KwOut{Distribution list $n_{col}$ and a set of resample points $X$;}
	
	\For{$i=1$ to $N_t$}{
		$ratio(i) \gets \omega_i*\mathcal{L}_{res}(T_i,\theta)/\sum_{i=1}^{N_t}{\omega_i*\mathcal{L}_{res}(T_i,\theta)}$\;
		$ num(i) \gets N_r*ratio(i)$\;
	}
	Gain distribution list $ n_{col} \gets num(i)$\;
	Generate LHS sampling points according to $n_{col}$ in each sub-domain and gather them in a set $X$ \;
	\KwRet{$n_{col},X$}\;
\end{algorithm}

The specific procedure is as follows. We utilize the temporal causal sampling distribution ratio as sampling distribution probability. Thus the number of sampled collocation points in $i$-th sub-domain is $N_r \times ratio(i)$. Then LHS is used to generate the corresponding number of random points. Based on Equation \eqref{rat}, when the temporal weight in $i$-th sub-domain is zero, none of the points are sampled in this sub-domain, which obeys temporal causality. For more general cases, for example, the temporal weight of $i$-th sub-domain is nonzero, then the number distribution of collocation points is based on the product of PDE residual loss and temporal weight. 

\begin{algorithm}[h]
	\caption{Adaptive Causal sampling method(ACSM)}\label{algorithm2}
	\label{alg:modifiedPINN}
	We use $N$ steps of gradient descent algorithm to update the parameters $\theta$:
	
	\For{$\epsilon=10^{-2},10^{-1},10^{0},10^{1},10^{2}$}{
		\For{$n=1$ to $N$}{
			Compute the temporal residual loss by		
			$\mathcal{L}_{res}(\theta)=\frac{1}{N_t}\sum_{i=1}^{N_t}\omega_i\mathcal{L}_{res}(T_i,\theta)$\;
			Compute the temporal causal weights by		
			$\omega_i=\exp(-\epsilon\sum_{k=1}^{i-1}\mathcal{L}_{res}(T_k,\theta)),for\ i=2,3,...,N_t$\;
			Resample $N_r$ collocation points by Algorithm\ref{alg:resample}\;
			Update the parameters $\theta$ via gradient descent\;
			$\theta_{n+1}=\theta_{n}-\eta\nabla_{\theta}\mathcal{L}_{res}(\theta_{n})$\;
		}
	}
\end{algorithm}

The pseudo-codes of our proposed resampling procedure and adaptive causal sampling method(ACSM) are given in Algorithm \ref{alg:resample} and \ref{alg:modifiedPINN} respectively. Here, we provide several remarks on these two algorithms. $N_r$ is a user-defined hyper-parameter that needs to be fixed before the training procedure. This hyper-parameter is problem-dependent and can be different in solving various PDEs. In this paper, we mainly discuss situations with few collocation points, which is a vital and efficient case. Besides, the extra resampling procedure will not increase computational complexity. In practice, we do the resampling procedure every 1000 iterations in the following numerical experiments.

\section{Numerical Experiments}\label{sec4}
In this section, we first illustrate the computational devices, reference solution and, error index, and then demonstrate some numerical experiments. 

We finish all the numerical experiments on a Nvidia GeForce RTX 3070 card. Besides, the reference solution are generated on a laptop with Intel(R) Core(TM) i7-10875H (2.30 GHz) and 16GB RAM. The software package used for all the computations is Pytorch 1.10. All the variables defined for computations in Pytorch are of float32 data type.

The reference solutions are all generated by the chebfun package \cite{chebfun}. The accuracy of prediction is calculated with the reference solution. Here we utilize the relative $L_2$ norm to evaluate the trained model performance:

\begin{equation}\label{error}
\begin{aligned}
\epsilon_{error}=\frac{\left[\tfrac{1}{N}\sum_{k=1}^{N}(\hat{u}(x_k,t_k)-u(x_k,t_k))^2\right]^{1/2}}{\left[\tfrac{1}{N}\sum_{k=1}^{N}(u(x_k,t_k))^2\right]^{1/2}},
\end{aligned}
\end{equation}
where $u(x_k,t_k)$ is the predicted solution and $\hat{u}(x_k,t_k)$ is the reference solution on a set of testing points $\{(x_k,t_k)\}_{k=1}^N, (x_k,t_k)\in\Omega\times\left(0,T\right]$.

We aim to demonstrate the effectiveness and efficiency of our proposed method in the following numerical experiments, where adaptive sampling suffers difficulties in converging to the accurate solution. Thus we can show that adaptive sampling should obey temporal causality and our proposed ACSM is a workable and efficient way. Due to the purpose of chasing the adaptive sampling method with limited points, the following numerical experiments are all set under the foundation that the total number of sampled points is unchanged and limited.

Before discussing the numerical results, we first introduce the comparison methods in the following numerical experiments: std-PINNs (original PINNs with fixed sampling \cite{PINN2019}), bc-PINN (backward compatible sequential PINN \cite{sequential}), CausalPINN-fixed (causal PINN \cite{Respectingcausality} with fixed sampling), CausalPINN-dynamic (causal PINN with dynamic sampling) and CausalPINN-adaptive (causal PINN with adaptive sampling). Besides, we select the well-performed and easy-implemented LHS\cite{LHS} as the basic sampling method. In this paper, all the fixed and dynamic sampling are based on LHS. The difference between fixed and dynamic is whether collocation points are resampled according to the number of training iterations.
More specifically, for fixed sampling, we utilize LHS only once at the beginning of training and fixed sampled collocation points during the whole training procedure. For dynamic sampling, we utilize LHS every 1000 training iterations to resample collocation points. For adaptive sampling, we utilize the adaptive sampling method with distribution ratio \eqref{adp}.

In the following numerical experiments, we mainly consider periodic boundary conditions, and hard constraint is utilized to strictly satisfy them. We embed the input coordinates into Fourier expansion by using
\begin{equation}\label{hard}
\begin{aligned}
v(x)=(1,\cos(\omega x),\sin(\omega x),\cos(2\omega x),\sin(2\omega x),...,\cos(m\omega x),\sin(m\omega x)),
\end{aligned}
\end{equation}
with $\omega=\frac{2\pi}{L}$ and m = 10. Here $L$ is the number of layers of the neural network. It can be proved that any $u_{\theta}(v(x))$ exactly satisfies a periodic constraint\cite{hard}. Then the loss function can be reduced to two parts
\begin{equation}\label{hardloss}
\begin{aligned}
L(\theta)=\lambda_{ic}\mathcal{L}_{ic}(\theta)+\lambda_{res}\mathcal{L}_{res}(\theta).
\end{aligned}
\end{equation}

\subsection{Cahn Hilliard Equation}\label{sec4.1}
We first consider the one-dimensional Cahn Hilliard equation, which is mentioned in Section \ref{sec2.2.2},
\begin{equation}\label{ch2}
\begin{aligned}
u_t-\nabla^2(r_2(u^3-u)-r_1\nabla^2{u})=0,& t\in \left (0,T\right],
\end{aligned}
\end{equation}
with the same initial and boundary conditions as Equation \eqref{ch}.

Cahn Hilliard equation plays an important role in studying diffusion separation and multi-phase flows. It has strong non-linearity and fourth-order derivative, which results in numerical challenges. Parameters in Equation \eqref{ch2} control the system evolution process. Specifically, the parameter $r_1$ represents the mobility parameter, and parameter $r_2$ is related to the surface tension at the interface. Different parameters lead to different physics evolution phenomena. 

To simplify the derivative calculation, we utilize the phase space representation which represents a high-order PDE into coupled with multiple lower-order PDEs. We introduce an intermediate function $\mu$:

\begin{equation}\label{mu}
\begin{aligned}
\mu=r_2(u^3-u)-r_1\nabla^2{u}.
\end{aligned}
\end{equation}

Then Equation \eqref{ch2} can be transformed to a system of equations

\begin{equation}\label{mueq}
\begin{aligned}
&u_t-\nabla^2(\mu)=0,\\
&\mu=r_2(u^3-u)-r_1\nabla^2{u}.
\end{aligned}
\end{equation}

As for the network architecture design, the two outputs of neural network are $\hat{u}_{\theta}(x,t)$ and $\hat{\mu}_{\theta}(x,t)$. Thus, the PDE residual loss can be divided into two parts, $\mathcal{L}_{res_1}(\theta)$ and $\mathcal{L}_{res_2}(\theta)$, namely
\begin{subequations}
	\begin{align}
	\mathcal{L}_{res_1}(\theta) &= \frac{1}{N_{r}}\sum_{i=1}^{N_{r}}|\hat{\mu}_{\theta}(t_{r}^i,x_{r}^i) - (r_2({\hat{u}}^3_{\theta}(t_{r}^i,x_{r}^i)-{u}_{\theta}(t_{r}^i,x_{r}^i))-r_1\nabla^2{{\hat{u}}_{\theta}}(t_{r}^i,x_{r}^i))|^2,\label{Za}\\
	\mathcal{L}_{res_2}(\theta) &= \frac{1}{N_{r}}\sum_{i=1}^{N_{r}}|\frac{\partial{{\hat{u}}_{\theta}}}{\partial{t}}(t_{r}^i,x_{r}^i) - \nabla^2{\hat{\mu}_{\theta}(t_{r}^i,x_{r}^i)}|^2\label{Zb}.
	\end{align}
\end{subequations}

Then the composite residual loss is
\begin{equation}\label{Zc}
\begin{aligned}
\mathcal{L}_{res}(\theta) &=\lambda_{res_1}\mathcal{L}_{res_1}(\theta)+\lambda_{res_2}\mathcal{L}_{res_2}(\theta).
\end{aligned}
\end{equation}

We also impose periodic boundary conditions as hard-constraints, thus the total training loss can be expressed as

\begin{equation}\label{wobc}
\begin{aligned}
\mathcal{L}(\theta)=\lambda_{ic}\mathcal{L}_{ic}(\theta)+\lambda_{res}\mathcal{L}_{res}(\theta).
\end{aligned}
\end{equation}

We choose $\lambda_{ic}=100$ and $\lambda_{res}=1$ to enforce the initial condition. Besides, to balance the magnitude of residual loss between Equation \eqref{Za}) and \eqref{Zb}, we choose $\lambda_{res1}=100$ and $\lambda_{res2}=1$.

\begin{figure}[htbp]
	\centering
	\includegraphics[width=1.0\textwidth]{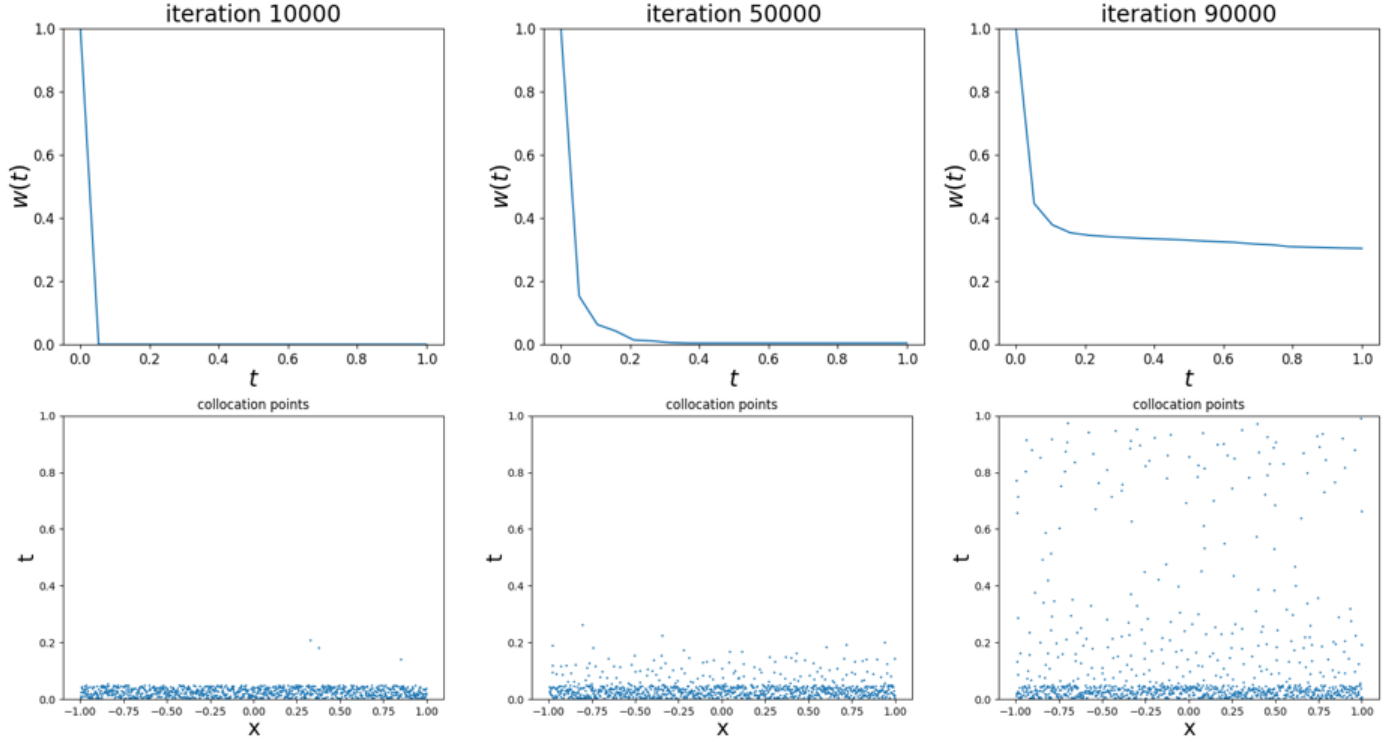}
	\caption{Cahn Hilliard equation(Case 1): Top: Temporal weights $w_t$ at different training iterations. Bottom: Distribution of collocation points in the whole spatio-temporal domain at corresponding iteration with Adaptive Causal Sampling.}
	\label{fig:ch2weightpoints}
\end{figure} 

\subsubsection{Case 1}
We consider a numerical experiment in paper \cite{sequential}. Let parameter $r_1=0.02$ and $r_2=1$. The entire computational field is $(x,t)\in\left[-1,1\right]\times\left[0,1\right]$. We divide the whole spatio-temporal domain into $n_t=20$ sub-domains evenly in time direction. Total 1000 collocation points
\begin{figure}[h!]
	\centering
	\includegraphics[width=1.0\textwidth]{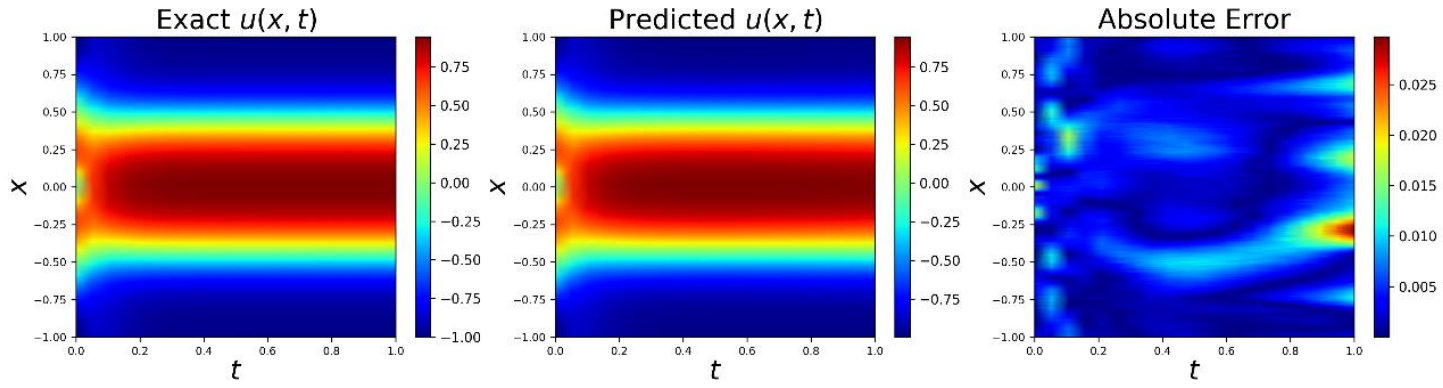}
	\caption{Cahn Hilliard equation(Case 1): Prediction and Error with Adaptive Causal Sampling with 1000 collocation points.}
	\label{fig:abserr}
\end{figure} 
are sampled in this case and the loss function is minimized by using $1\times 10^5$ ADAM iterations. The network architecture used has 4 hidden layers with 128 neurons in each layer and the output layer has two neurons, namely $\hat{u}_{\theta}(x,t)$ and $\hat{\mu}_{\theta}(x,t)$. A Runge-Kutta time integrator with time step $\Delta t = 10^{-5}$ is used in chebfun.

\begin{figure}[htbp]
	\centering
	\includegraphics[width=1.0\textwidth]{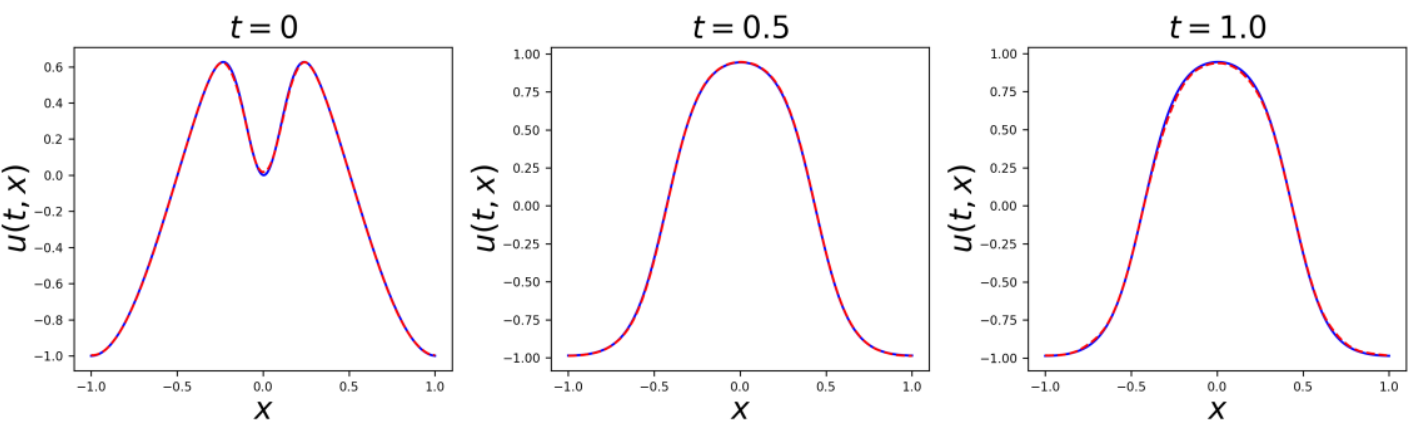}
	\caption{Cahn Hilliard equation(Case 1): Result Snapshots with Adaptive Causal Sampling with 1000 collocation points.}
	\label{fig:snap}
\end{figure} 

\begin{table}[h!]
	\centering
	\begin{tabular}{ccccc}
		\toprule
		\multicolumn{1}{c}{Method} &\multicolumn{1}{c}{collocation points}  &\multicolumn{1}{c}{Relative $L^2$ error} \\
		\hline
		\multicolumn{1}{c}{std-PINNs}& $2\times10^6$ & 8.594$e^{-1}$\cite{sequential}\\
		\multicolumn{1}{c}{bc-PINN}& $2\times10^6$ & 1.86$e^{-2}$\cite{sequential}\\
		\multicolumn{1}{c}{CausalPINN-fixed}& 1000 & 4.796$e^{-1}$\\
		\multicolumn{1}{c}{CausalPINN-dynamic}& 1000 & 4.632$e^{-1}$\\
		\multicolumn{1}{c}{CausalPINN-adaptive}& 1000 & 4.684$e^{-1}$\\
		\multicolumn{1}{c}{ACSM}& 1000 & \pmb{7.046$e^{-3}$}\\
		\bottomrule
	\end{tabular}
	\caption{Cahn Hilliard equation(Case 1): Relative $L^2$ errors obtained by different sampling methods.}
	\label{tab:1}
\end{table}

The result of temporal causal weight and sampled points is in Fig \ref{fig:ch2weightpoints}. We can see that as the iteration increases, collocation points are gradually sampled from the initial time to the later one, which respects the temporal causality. The predict results with our proposed method are shown in Fig \ref{fig:abserr} and \ref{fig:snap}. Compared with adaptive sampling in the illustrative example of Section \ref{sec2.2.2}, our method obeys temporal causality and converges to the accurate solution, which proves the effectiveness of ACSM. As for the relative $L_2$ error, ours is $7.046E-03$ in this case which is two magnitudes less than that without causal sampling $4.684E-01$. More comparison details of different sampling methods are shown in Table \ref{tab:1}. ACSM achieves the best performance among these sampling methods.


\subsubsection{Case 2}
Then we change the value of parameters, and set $r_1=0.01$ and $r_2=1$, which weakening the mobility effect. Computational field is $(x,t)\in\left[-1,1\right]\times\left[0,0.25\right]$. In this case, we divide the whole spatio-temporal domain into $n_t=50$ sub-domains evenly in time direction. Total 3000 collocation points in this experiment and the loss function is minimized by using $1\times 10^5$ ADAM iterations. The network architecture is the same as that in case 1. 

 \begin{figure}[h!]
	\centering
	\includegraphics[width=1.0\textwidth]{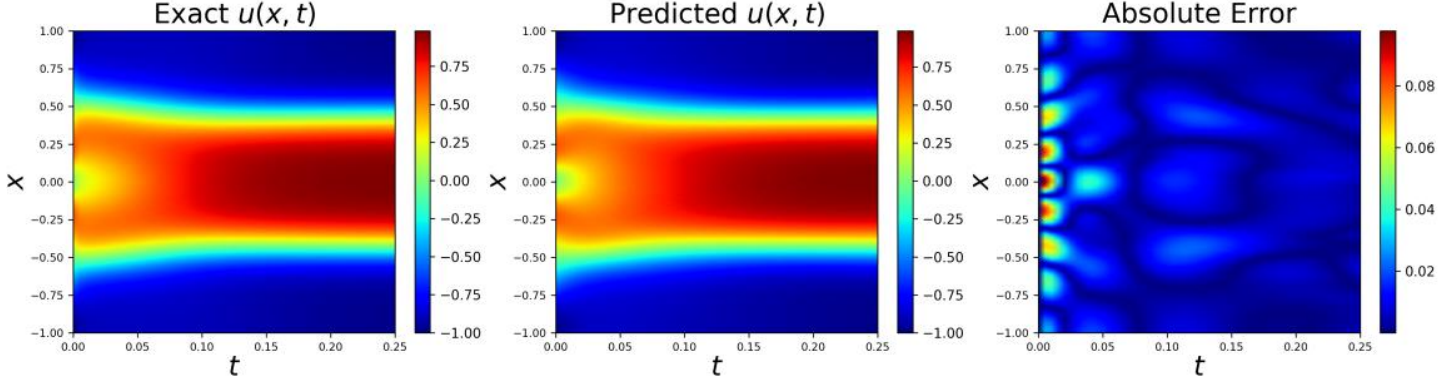}
	\caption{Cahn Hilliard equation(Case 2): Prediction and Error with Adaptive Causal Sampling with 3000 collocation points.}
	\label{fig:ch4abserr}
\end{figure}

\begin{figure}[h!]
	\centering
	\includegraphics[width=1.0\textwidth]{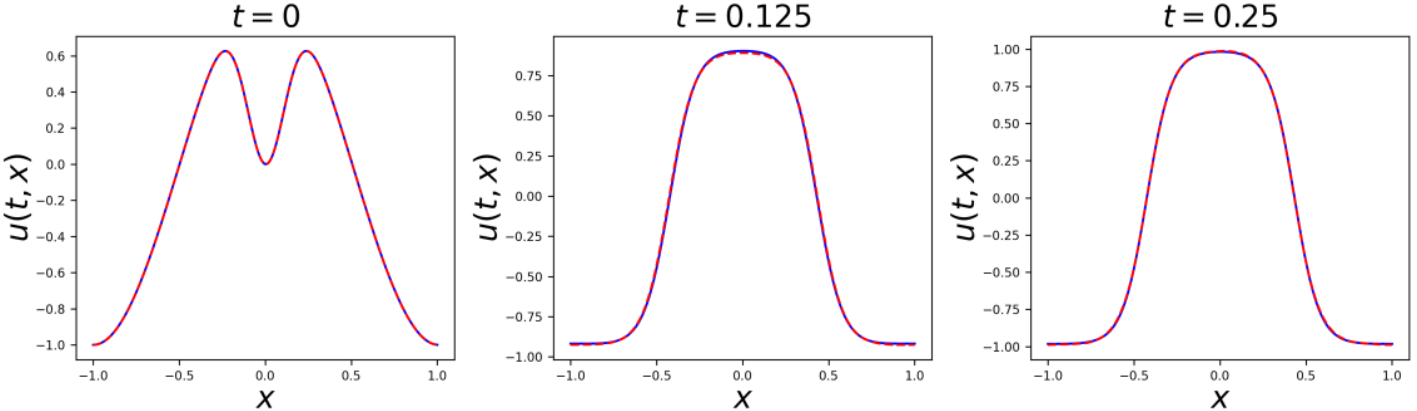}
	\caption{Cahn Hilliard equation(Case 2): Result Snapshots with Adaptive Causal Sampling with 3000 collocation points.}
	\label{fig:ch4snap}
\end{figure}

In this case, the solution changes more violently at the beginning of physics evolution than the latter one, which brings more challenges in numerical modeling. Fig \ref{fig:ch4abserr} and \ref{fig:ch4snap} show the prediction results. We gain the relative $L^2$ error $1.633E-02$ in this case. Fig \ref{fig:ch2weightpoints} shows that the proposed sampling method obeys temporal causality and collocation points are gradually sampled in latter sub-domains as training iteration increases.
 
\begin{table}[htb]
	\centering
	\begin{tabular}{ccccc}
		\toprule
		\multicolumn{1}{c}{Method} &\multicolumn{1}{c}{collocation points}  &\multicolumn{1}{c}{Relative $L^2$ error} \\
		\hline
		\multicolumn{1}{c}{std-PINNs}& 3000 & 2.213$e^{-2}$\\
		\multicolumn{1}{c}{CausalPINN-fixed}& 3000 & 4.452$e^{-1}$\\
		\multicolumn{1}{c}{CausalPINN-dynamic}& 3000 & 5.815$e^{-2}$\\
		\multicolumn{1}{c}{CausalPINN-adaptive}& 3000 & 4.076$e^{-2}$\\
		\multicolumn{1}{c}{ACSM}& 3000 & \pmb{1.633$e^{-2}$}\\
		\bottomrule
	\end{tabular}
	\caption{Cahn Hilliard equation(Case 2): Relative $L^2$ errors obtained by different sampling methods.}
	\label{tab:2}
\end{table}

\subsection{Korteweg-de Vries (KdV) Equation}\label{sec4.2}
We consider the Korteweg–de Vries (KdV) equation, which describes the shallow water surfaces that interact weakly and nonlinearly, and long inner waves in densely layered oceans. The one-dimensional KdV equation takes the form
\begin{equation}
\begin{aligned}
\frac{\partial u}{\partial t} + \lambda_1 u\frac{\partial u}{\partial x} + \lambda_2 \frac{\partial^3 u}{\partial x^3}&=0, x \in (0,1),t \in (0,1],\\
u_0(x) &= u(x,t=0).
\end{aligned}
\end{equation}
with parameters $(\lambda_1,\lambda_2)$. Here, we choose parameters as $(\lambda_1=1,\lambda_2=0.0025)$ from \cite{PINN2019}. KdV equation also contains high-order derivatives.

The initial condition is $u_0(x)=cos(\pi x)$ and boundary conditions are periodic. To gain the training and test data, we use Chebfun package to simulate the KdV equation up to a final time $t=0.8$ with time-step $\Delta t=10^{-6}$. The entire computational domain is $(x,t)\in\left[-1,1\right]\times\left[0,0.8\right]$. The PDE residual loss is defined as:
\begin{equation}\label{losskdv}
\begin{aligned}
\mathcal{L}_{res}(\theta)= \frac{1}{N_{r}}\sum_{i=1}^{N_{r}}|\frac{\partial{{\hat{u}}_{\theta}}}{\partial{t}}(t_{r}^i,x_{r}^i) + {\hat{u}_{\theta}}\frac{\partial{{\hat{u}}_{\theta}}}{\partial{x}}(t_{r}^i,x_{r}^i) + 0.0025\frac{\partial^3 {{\hat{u}}_{\theta}}}{\partial {x^3}}(t_{r}^i,x_{r}^i)|^2.
\end{aligned}
\end{equation}

We also impose periodic boundary conditions as hard-constraints, thus the aggregate training loss can be expressed as

\begin{equation}
\begin{aligned}
\mathcal{L}(\theta)=\lambda_{ic}\mathcal{L}_{ic}(\theta)+\lambda_{res}\mathcal{L}_{res}(\theta).
\end{aligned}
\end{equation}

\begin{figure}[htbp]
	\centering
	\includegraphics[width=1.0\textwidth]{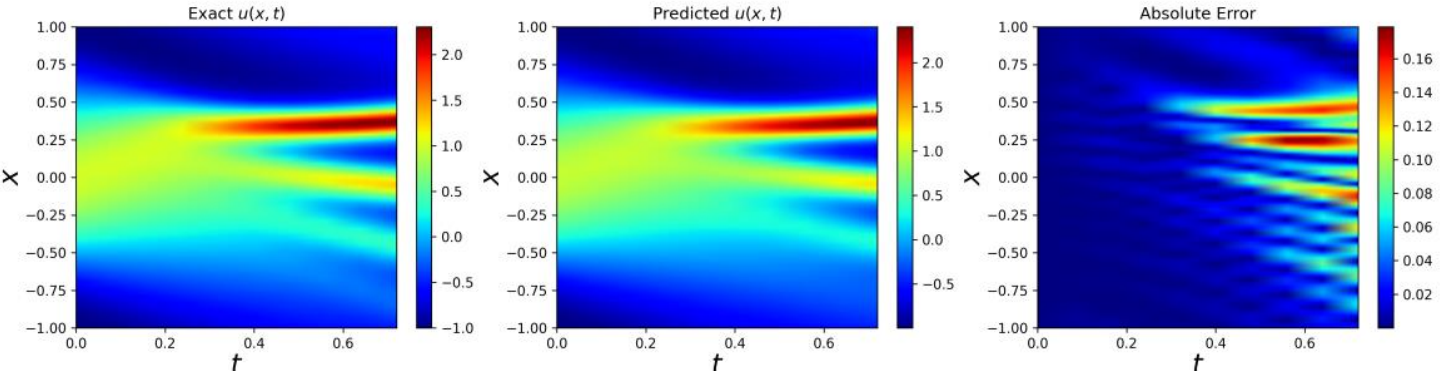}
	\caption{KdV equation: Prediction and Error with Adaptive Causal Sampling with 300 collocation points.}
	\label{fig:kdvabserr}
\end{figure} 

\begin{figure}[h!]
	\centering
	\includegraphics[width=1.0\textwidth]{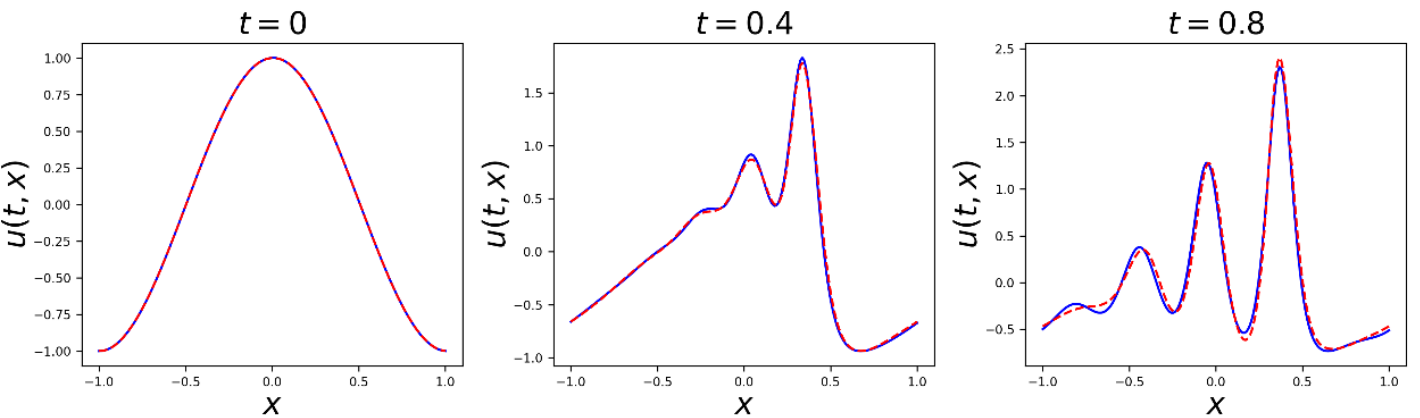}
	\caption{KdV equation: Result Snapshots with Adaptive Causal Sampling with 300 collocation points.}
	\label{fig:kdvsnap}
\end{figure} 

We choose $\lambda_{ic}=100$ and $\lambda_{res}=1$ to enforce the initial condition. 
Besides, We divide the whole spatio-temporal domain into $n_t=10$ sub-domains evenly in time direction. We use total 300 collocation points in this experiment and the loss function is minimized by using $1\times 10^5$ ADAM iterations. The network architecture used has 4 hidden layers with 128 neurons in each layer and the output layer has one neuron $\hat{u}_{\theta}(x,t)$. 

Figure \ref{fig:kdvabserr} and \ref{fig:kdvsnap} show the prediction results of our proposed methods. Besides, the comparison results between different sampling methods are shown in Table \ref{tab:3}, which illustrates that ACSM achieves best with very few collocation points.


\begin{table}[h!]
	\centering
	\begin{tabular}{ccccc}
		\toprule
		\multicolumn{1}{c}{Method} &\multicolumn{1}{c}{collocation points}  &\multicolumn{1}{c}{Relative $L^2$ error} \\
		\hline
		\multicolumn{1}{c}{std-PINNs}& 300 & 5.348$e^{-1}$\\
		\multicolumn{1}{c}{CausalPINN-fixed}& 300 & 5.966$e^{-1}$\\
		\multicolumn{1}{c}{CausalPINN-dynamic}& 300 & 3.703$e^{-1}$\\
		\multicolumn{1}{c}{CausalPINN-adaptive}& 300 & 1.335$e^{-1}$\\
		\multicolumn{1}{c}{ACSM}& 300 & \pmb{5.662$e^{-2}$}\\
		\bottomrule
	\end{tabular}
	\caption{KdV equation: Relative $L^2$ errors obtained by different sampling methods.}
	\label{tab:3}
\end{table}

\subsection{Discussion of Sampling Efficiency}\label{sec4.3}

\begin{figure}[h]
	\centering
	\includegraphics[width=1.0\textwidth]{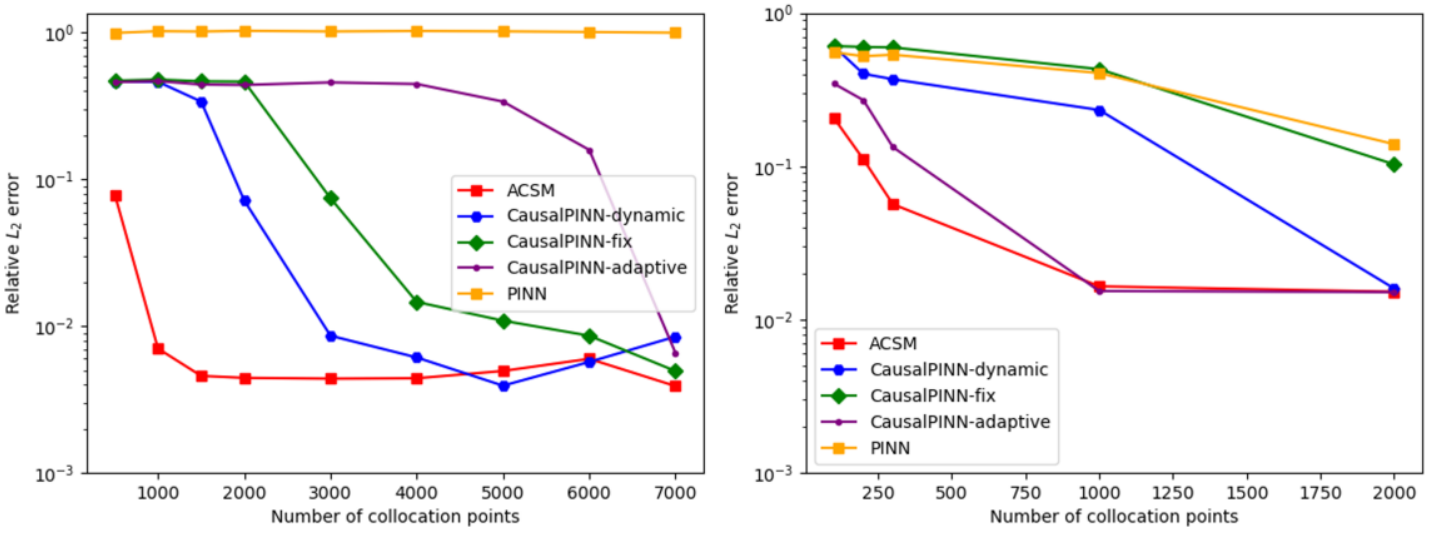}
	\caption{Sampling efficiency of different methods. Left: Cahn Hilliard equation (case 1). Right: KdV equation.}
	\label{fig:samplingefficiency}
\end{figure} 

In subsection \ref{sec4.1} and \ref{sec4.2}, we explore the prediction performance of our proposed method. In this subsection, we aim to discuss the sampling efficiency of different sampling methods. Here we show two efficiency experiments in Cahn Hilliard equation and KdV equation respectively. Figure \ref{fig:samplingefficiency} shows that when the number of collocation points is small, our proposed method performs better. Specifically, in the left sub-figure of Figure \ref{fig:samplingefficiency}, ACSM can achieve $10^{-3}$ magnitude of relative $L_2$ error in solving Cahn Hilliard equation (case 1), however other comparison methods only achieve $10^{-1}$ or $10^{0}$ magnitude, which shows the superiority of ACSM in few points scenario. When the number is large, the relative $L_2$ error of ACSM is also comparable with other methods and stays the same error magnitude. In the right sub-figure, we can get a similar conclusion that ACSM performs better than others, especially in a few points scenario, and brings almost no extra computational cost, which demonstrates the sampling efficiency of ACSM. Due to such sampling efficiency in fewer data scenarios, our proposed method shows potential in solving high-dimensional and computationally complex PDEs, which always suffers from expensive computational cost and heavy time consumption.


\section{Conclusion}\label{sec5}
In this paper, we focus on the performance of sampling methods for PINNs in solving PDEs. First, we analyze the failure of adaptive sampling method in few data scenario and bring out the argument that sampling should obey temporal causality. An adaptive causal sampling method for PINNs is proposed, which obeys the temporal causality. Numerical experiments are investigated on several nonlinear PDEs with high-order derivatives and strong nonlinearity, including Cahn Hilliard and KdV equations. However, the method in general can be applicable to other PDEs. By comparisons between different sampling methods, we conclude that our proposed adaptive causal sampling method improves the prediction performance up to two orders of magnitude and increases computational efficiency of PINNs with few collocation points, which shows potential in large-scale problems and high-dimensional problems.

\section{Acknowledgments}
The author thank Ph.D. candidate Ziyuan Liu and master candidate Kaijun Bao for their advice in revision of this paper. This work was partially supported by the Key NSF of China under Grant No. 62136005, the NSF of China under Grant No. 61922087, Grant No.61906201 and Grant No. 62006238. 


\bibliographystyle{elsarticle-num}
\bibliography{causal}

\end{document}